%% file: main.tex
\pdfoutput=1
\documentclass[11pt]{article}

\usepackage{acl}
\usepackage{graphicx} 
\usepackage{times}
\usepackage{latexsym}
\usepackage{booktabs}
\usepackage{enumitem}
\usepackage{multirow}
\usepackage{array}
\usepackage[T1]{fontenc}
\usepackage[utf8]{inputenc}
\usepackage{todonotes}

\usepackage{microtype}
\usepackage{float}
\usepackage[capitalise]{cleveref}
\usepackage{siunitx}
\usepackage{colortbl}
\definecolor{lg}{gray}{0.89}

\newcommand{\kmeans}[0]{$K$-means}

\title{Why Did the Chicken Cross the Road?\\ Rephrasing and Analyzing Ambiguous Questions in VQA}

\author{Elias Stengel-Eskin\hspace{1.5em} Jimena Guallar-Blasco\hspace{1.5em} Yi Zhou \hspace{1.5em} Benjamin Van Durme\\
Johns Hopkins University\\
\texttt{\{elias, jgualla1, yzhou188, vandurme\}@jhu.edu}}
\date{}

\begin{document}
\maketitle
\begin{abstract}
Natural language is ambiguous. Resolving ambiguous questions is key to successfully answering them.
%Resolving ambiguities in questions is key to successfully answering them.
Focusing on questions about images, we create a dataset of ambiguous examples. We annotate these, grouping answers by the underlying question they address and rephrasing the question for each group to reduce ambiguity. 
Our analysis reveals a linguistically-aligned ontology of reasons for ambiguity in visual questions. 
We then develop an English question-generation model which we demonstrate via automatic and human evaluation produces less ambiguous questions. 
We further show that the question generation objective we use allows the model to integrate answer group information without any direct supervision.\footnote{Code and data: \url{https://github.com/esteng/ambiguous_vqa}}
\end{abstract}

\section{Introduction}
The ability to ask questions allows people to efficiently fill knowledge gaps and convey requests; this makes questions a natural interface for interacting with digital agents.  
Visual question answering (VQA) models more specifically seek to answer questions about images, which can be useful in a variety of settings, such as assistive tech \citep{bigham.j.2010}.  
A number of datasets have been proposed for training VQA models, including VQAv2 \citep{goyal.y.2017}, VizWiz \citep{gurari.d.2018}, and GQA \citep{hudson.d.2019}. 
Such datasets are not only useful for training -- they represent the aggregate judgements of speakers on a variety of factors, including ambiguity. 

\begin{figure}[t]
    \includegraphics[width=\linewidth]{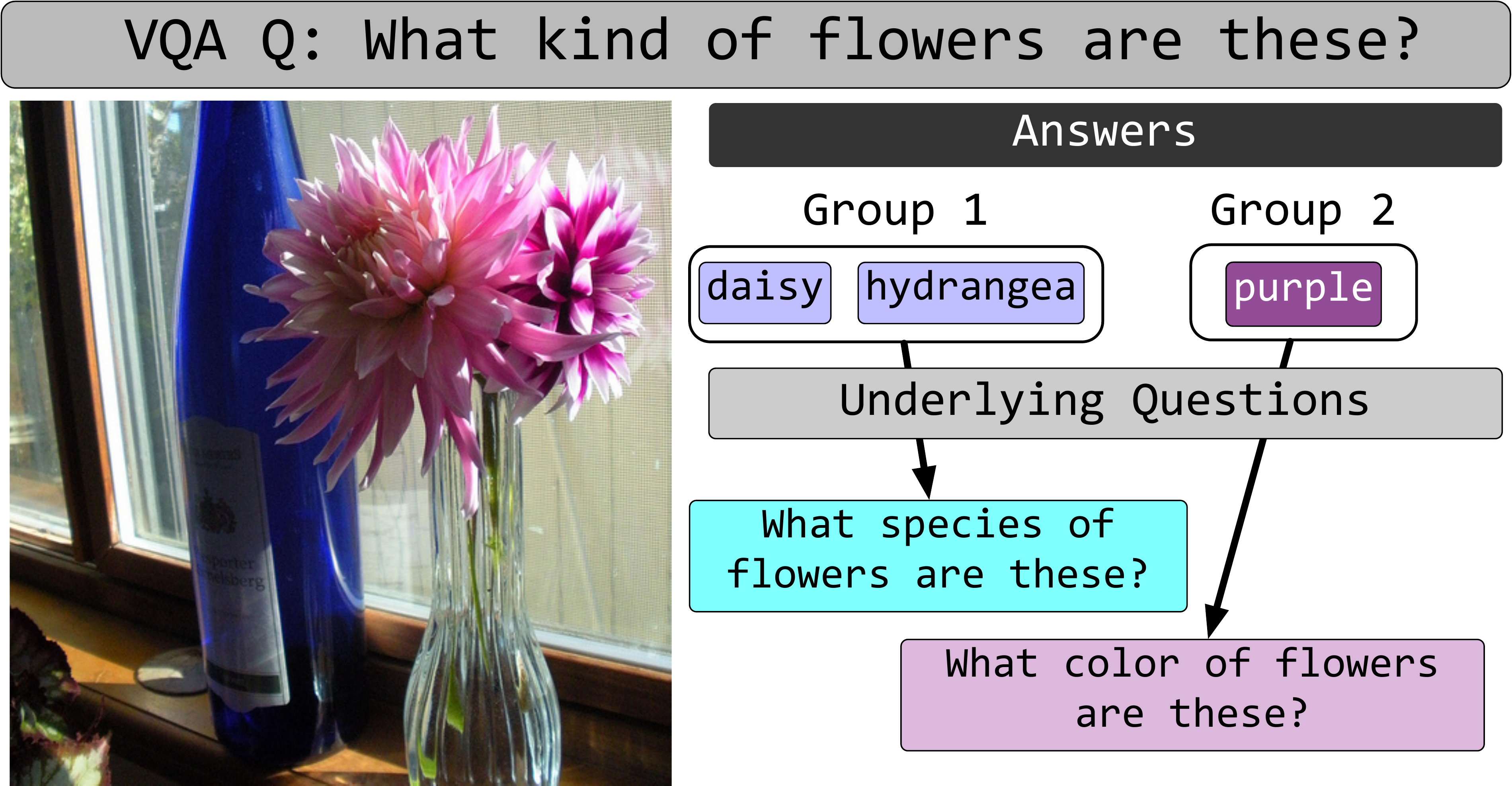}
    \vspace{-1em}
    \caption{An ambiguous visual question from our dataset. Answers are grouped by the underlying question they answer, and the question is rephrased for each group. 
    Answers within a group do not necessarily match, but do answer the same question.}
    \label{fig:fig1}
    \vspace{-1em}
\end{figure}

Ambiguity is a core feature of natural language, and can exist at all levels of linguistic analysis \citep{piantadosi.s.2012}. 
In the context of data annotation, ambiguity often leads to disagreement between annotators.
Given that the data resulting from crowdsourced annotation projects is typically used in a categorical fashion to train and evaluate models, annotator disagreements are problematic.
Past work has often looked at detecting and resolving disagreements from the perspective of trust \citep{hovy.d.2013}, where some annotators are assumed to be more or less trustworthy.
However, in the case of ambiguity, an annotator's honest effort might still lead to disagreement; in such cases, collecting more annotations can fail to establish a consensus.
This differs from disagreements arising as a result of mistakes and cheating, where gathering more annotations would effectively outvote low-quality annotations.
Ambiguity in the context of questions presents a particularly rich problem:
firstly, from a formal point of view, question semantics are an area of active development and debate \citep{ginzburg.j.2010}; this makes empirical accounts of questions particularly useful. 
Secondly, questions are increasingly relevant to natural language processing (NLP) research.
Many NLP tasks are cast as question-answering (QA), including a growing number of tasks which can be cast as few-shot QA. 
Against this backdrop, we aim to document and describe ambiguities in VQA as well as to introduce a model for resolving them.

Our main contributions are: 
\begin{enumerate}[noitemsep,nolistsep,topsep=0pt]
\item We examine how ambiguity appears in the VQAv2 data by constructing a dataset of 1,820 annotated visual image-question-answer triples.  For each question, we ask annotators to re-group answers according to the underlying question they answer, and to rewrite questions to unambiguously correspond to that group. 
An example from our dataset can be seen in \cref{fig:fig1}. 
For the ambiguous VQA question given at the top, annotators group existing answers into two topical groups (\emph{species} and \emph{color}).
Then, for each group, annotators rewrite the original question such that it could be answered by answers from the corresponding group, but not from other groups. 
\item We create an ontology of causes for linguistic ambiguity based on the PropBank ontology \citep{kingsbury.p.2002, gildea.d.2002, palmer.m.2005}, and annotate our data with these causes. 
\item We develop a visual question generation model which learns to rewrite questions; we validate this model with the re-grouped answers and re-written questions from our dataset. 
Our model can be used to cluster answers into their groups without any supervision for answer groups. 
\end{enumerate}

\section{Ambiguity in VQA}\label{sec:ambiguity} 
In the VQAv2 annotations, each image has multiple questions,
with each question being redundantly answered by up to 10 annotators.
This redundancy is crucial for our annotations, as it provides us with multiple judgments per question, some of which may indicate ambiguity. 
We define ambiguous examples as ones where annotators are responding to different underlying questions.
Note that this definition is not exhaustive, as it relies on the annotations; an example could be ambiguous but have few annotations, resulting in complete agreement between annotators.
We contrast this definition with visual underspecification and uncertainty, which are categorized by a lack of visual information needed to answer a question, rather than ambiguity about what the question is.
These can appear simultaneously, e.g. in \cref{fig:underspec_ambig} where there is both ambiguity and underspecification. 
\begin{figure}[t]
    \centering
    \includegraphics[width=0.47\textwidth]{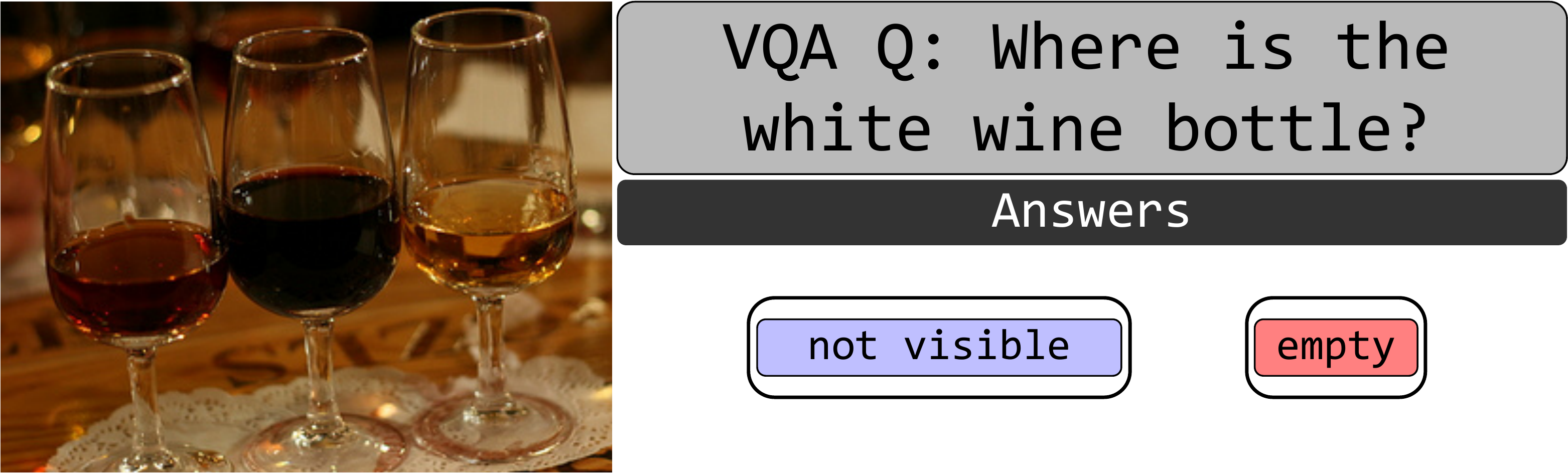}
    % \vspace{-0.5em}
    \caption{A visually underspecified question.}
    \label{fig:underspec}
    \vspace{-0.5em}
\end{figure}

\cref{fig:underspec} gives an example of underspecification, as the information being queried is absent in the image and must be inferred. 
Past efforts examining reasons for annotator disagreement in VQA have addressed this distinction: \citet{bhattacharya.n.2019} introduce a dataset of 45,000 VQA examples annotated with reasons for disagreement, including ambiguity and lack of visual evidence as two separate categories. 
In practice, however, many examples labeled as ambiguous (such as \cref{fig:underspec}) are cases of underspecification or unambiguous questions paired with visually ambiguous images.
We use the ambiguous examples from \citet{bhattacharya.n.2019} as a starting point for our dataset. 
\begin{figure}[t]
    \centering
    \includegraphics[width=0.47\textwidth]{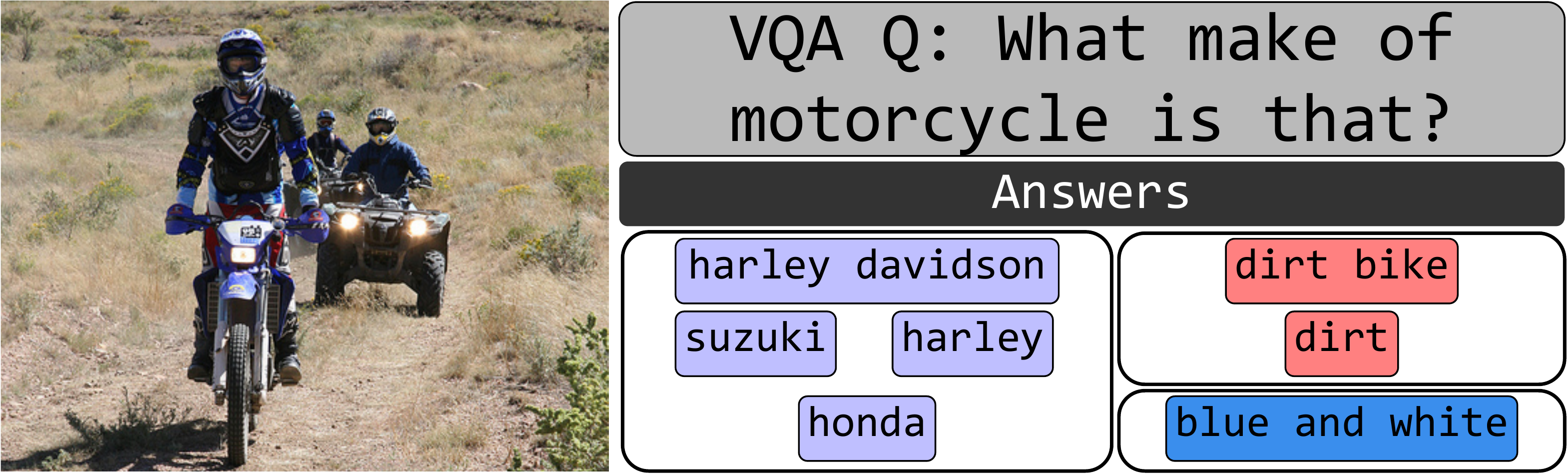}
    % \vspace{-0.5em}
    \caption{An underspecified and ambiguous question.}
    \label{fig:underspec_ambig}
    \vspace{-0.5em}
\end{figure}
\section{Data} \label{sec:data} 
To properly study linguistic ambiguity in VQA, we collect a dataset of ambiguous examples, which represents a resource for categorizing and analyzing ambiguous questions and contains 1,820 answers to 241 image-question pairs. 
The data contains answers grouped by their underlying questions; there are 629 rewritten underlying questions.
Our dataset is intended for evaluating models and for performing analysis, not for training.

The size of the ambiguous subset of VQA from \citet{bhattacharya.n.2019} prohibits our re-annotating the whole dataset, so we create a subset of data that is likely to be linguistically ambiguous.
First, we sort the annotations into a priority queue using several heuristics. 
To merge synonymous answers (e.g. ``cat'', ``the cat'', ``feline'') we embed each answer into continuous space using GloVe embeddings \citep{pennington.j.2014}, mean-pooling across words for multi-word answers and apply \kmeans{}  \citep{macqueen.j.1967, lloyd.s.1982} to the resulting embeddings, iteratively increasing the number of clusters $k$.
Examples are scored by combining the \kmeans{} inertia score with a penalty for each additional cluster, trading off cluster coherence and having as few clusters as possible. 
These are subsequently sorted by how balanced their clusters are -- balanced clusters are more likely to be ambiguous, as unbalanced clusters are often a result of a single bad annotation.
We remove yes-no questions with only ``yes'' and ``no'' answers, as they answer the same question.
Note that we do not include questions from GQA \citep{hudson.d.2019} in our dataset.
Because GQA questions were generated from a grammar rather  by annotators, we do not expect there to be as much ambiguity. 
Furthermore, GQA questions are not part of the labeled data from \citet{bhattacharya.n.2019}.

\input{ontology_table}
% \vspace{-0.5em}
\subsection{Annotation Interface}
We introduce a new annotation interface for re-grouping answers and re-writing questions (cf. \cref{append:interface}). 
We present the annotators with the question, image, and answers; answers are pre-grouped based on the GLoVe \kmeans{} cluster assignments and are drag-able. 
Each answer cluster is paired with an editable text-box containing the original question. 
For each example, annotators have 3 tasks: first, they must decide whether the answers provided in the example correspond to different questions, or whether they all answer the same underlying question, i.e. whether the question is ambiguous or not. 
If an example is identified as being ambiguous, the second task is to re-group annotations by the question they answer.
Each answer can be dragged into the appropriate cluster or deleted if it is spam; new clusters can also be created, and empty clusters can be deleted.
Annotators were instructed to cluster answers by their underlying question, \emph{not} by whether they are semantically similar.
For example, antonyms like ``good'' and ``bad'' may be grouped into the same answer cluster. 
Finally, in the third task, annotators were asked to minimally edit the question corresponding to each created cluster, such that the new question uniquely corresponds to that cluster of answers.
Instructions were presented to the annotators in text and video format. 
A \emph{local pilot} with two vetted annotators was run to collect data for filtering annotators on Amazon MechanicalTurk (MTurk).
A further \emph{MTurk pilot} was run and only annotators with high agreement to the local annotators were allowed to participate in further annotation.
Note that the local annotators were unfamiliar with the goals of the project (i.e. not the authors) and paid by time, not by number of annotations.  
See \cref{append:ontology} for details on the crowdsourcing process, including wage information.
At least one author manually vetted all ambiguous examples, discarding noisy examples and editing questions for fluency. Examples were eliminated if the question could not be answered from the corresponding image (e.g. \cref{fig:underspec}), or if the image had one or fewer viable responses. Edited questions were changed only to improve the grammaticality of the rephrased questions; their content was left unedited.

% \vspace{-0.5em}
\subsection{Statistics} 
Of the 1,249 examples run through MTurk, annotators skipped 942, identifying $307$ as ambiguous. 
After cleaning these examples we have 241 unique image-question combinations, corresponding to 629 unique rewritten questions (including the examples from the pilot.) 
Each rewritten question is paired with $1$-$9$ unique answers (mean: 2.9) -- note that questions can have only one answer, since each example has multiple rewritten questions.
We split our data into $30$ dev questions and $211$ test questions. 

% \vspace{-0.5em}
\subsection{Inter-annotator Agreement}
We measure agreement on two levels: to what extent annotators identified the same examples as ambiguous, and the overlap between clusters of answers.
Note that perfect inter-annotator agreement cannot be expected. 
Given that the examples we are interested in were ambiguous to the original set of VQAv2 annotators, with some seeing one reading over another, it is likely that some of the annotators in our task would also see only one reading.  

\emph{Ambiguity agreement} is defined as the percentage of examples two annotators both marked as being ambiguous. 
This number is averaged across annotator pairs. 
In the local pilot, the annotators had a pairwise ambiguity agreement score of  $79.5\%$.
In the MTurk pilot, 5 annotators had a mean pairwise score of $73.5\%$ with a standard deviation of $6.0\%$ (min $62.5\%$, max $80.0\%$). 
Note that we obtained redundant annotations only for the local and MTurk pilot HITs, and not the main data collection HIT.

The \emph{cluster agreement} between two annotators is defined as the F1 score between the clusters of answers produced. 
Since the clusters are not aligned a priori, we use the Hungarian algorithm \citep{kuhn.h.1955} to find a maximum overlap bipartite matching between clusters from each annotator and then compute the F1 score between aligned clusters. 
These scores are averaged across annotator pairs.
The local pilot cluster agreement score was $92.2$, and the MTurk pilot's score was $88.4$, with a standard deviation of $6.0$ (min $77.1$, max $94.6\%$). 

% \vspace{-0.5em}
\subsection{Ambiguity Ontology} \label{sec:data_ontology}
After collecting the data, we observed that there were multiple groups within the ambiguous examples, corresponding to the factors that made a question ambiguous. 
We manually annotated all ambiguous examples according to the following linguistically-grounded ontology, which is largely aligned to PropBank roles \citep{kingsbury.p.2002, gildea.d.2002, palmer.m.2005}. 
The ontology is divided broadly into 3 categories. Property-based questions typically have to do with objects with multiple properties, and relate to partition question semantics \citep{groenendijk.j.1984}; more information can be found in \cref{sec:groenendijk}. Dynamic questions are about dynamic properties of objects or events. Finally, pragmatic ambiguities mainly relate to ambiguity in inferring the intention of the questioner, including choosing which element of the the world is most salient. 
Each category contains several sub-categories -- these are summarized in \cref{tab:ontology} and described in-depth in \cref{append:ontology}.

\begin{figure}
    \centering
    \includegraphics[width=\linewidth]{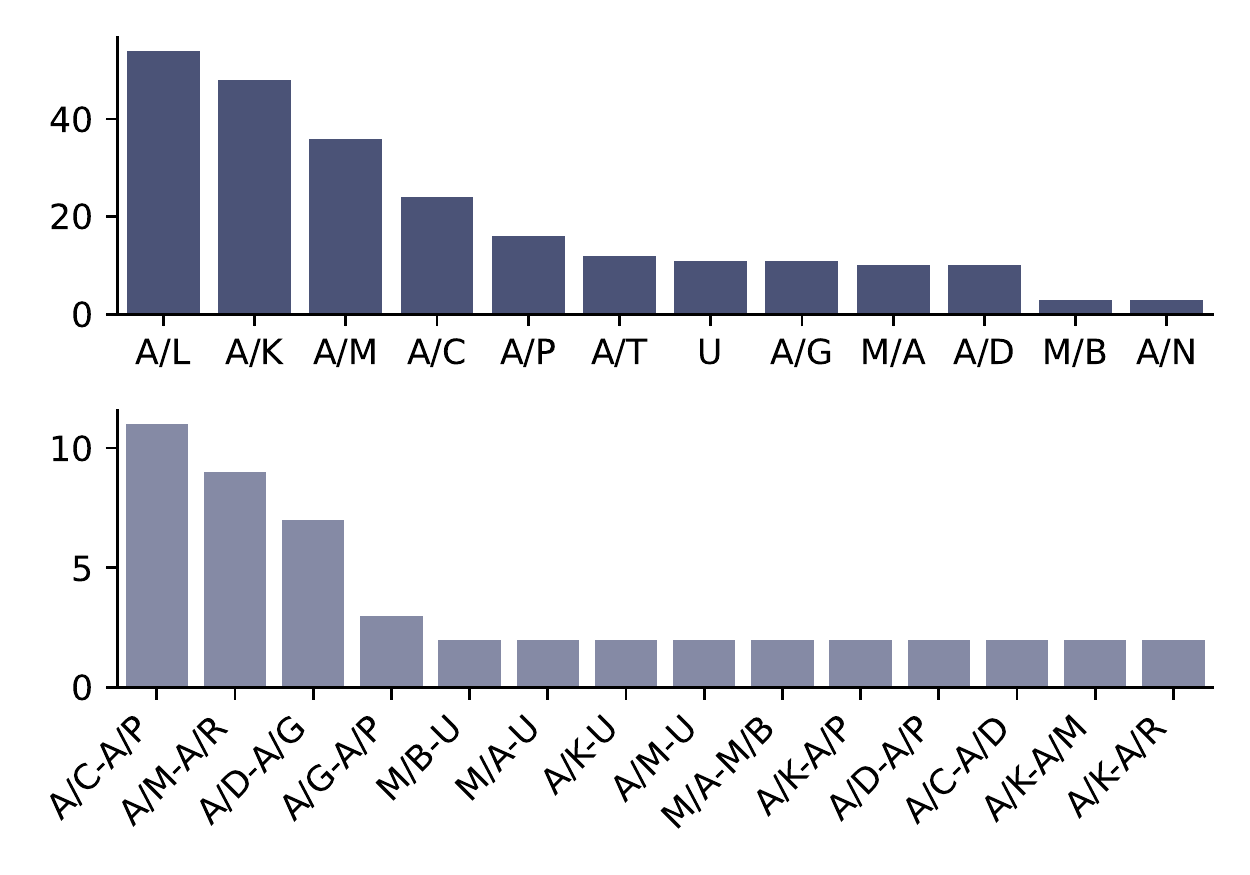}
    \vspace{-2em}
    \caption{(Top) Frequency of each category. (Bottom) Co-occurrence frequency of each category (excluding frequencies $\leq 1$). Some categories are highly correlated, indicating higher-order ambiguity.}
    \vspace{-1em}
    \label{fig:cat_freq}
\end{figure}

\cref{fig:cat_freq} shows the frequency of each category in our data, with the most common categories being location, kind, and multiple options, and shows the frequency with which pairs of categories co-occur (excluding pairs that only co-occur once). 
Several categories co-occur frequently, indicating higher-order ambiguity (i.e. ambiguity between what type of question is being asked). 
For example cause and purpose often co-occur; this indicates that they are often confused for each other, with some annotators providing answers consistent with a cause interpretation and others with a purpose interpretation. 
Furthermore, that they do not always co-occur indicates that ambiguity exists even within one interpretation. 

\section{Model}
The data collected in \cref{sec:data} consists of questions rewritten according to their answer clusters.
We develop a visual question generation (VQG) model which takes in answers and images and produces questions.
After confirming the performance of the VQG model for generation generally, we evaluate the performance of a VQG model with respect to the answer clusters in our dataset.
Specifically, we examine how the model can be used for clustering answers within an answer group together. 
Given that the answer clusters are based on the underlying question the answer is answering, we hypothesize that a good VQG model should not only learn to generate questions with a high similarity to the reference questions, but learn input representations that contain answer group information. 
Note that this information would have to emerge in an unsupervised fashion, as we do not provide any answer group information during training. 

We present a simple model for VQG consisting of a pre-trained vision-language  encoder followed by a pretrained text-to-text encoder-decoder model. 
The encoder embeds an image and an answer into a shared representation space; the decoder produces a text question conditioned on this shared representation. 
We use ViLT \citep{kim.w.2021} as our vision-language encoder.
ViLT is a pre-trained fully transformer-based 87.4M-parameter model.
The available ViLT model fine-tuned for VQA was trained on the entirety of the VQAv2 training data; since the annotations for \citet{bhattacharya.n.2019} come from the training set, our annotations also are sourced from the VQAv2 training set. 
To avoid test-set leakage, we fine-tune our own version of ViLT on a modified training set that excludes our annotations.
Our input to ViLT is the image $I_i$ and a text answer $a_i$ from the set of answers for instance $i$, $A_i$.
To generate text, we feed the output of ViLT to a pre-trained T5-base encoder-decoder model \citep{raffel.c.2020} with $~\sim 220$M parameters, accessed via Huggingface Transformers \citep{wolf.t.2020}. 
We replace the T5 embedding layer with the output of our ViLT encoder, and train the model using all answers in the dataset with ``yes'' or ``maybe'' confidence ratings.
We use categorical cross-entropy loss computed against the original question $Q_i$ as our loss function.
Note that the question $Q_i$ is taken directly from the VQAv2 data, which we refer to as ``original data'' -- we do not train on the annotations collected in \cref{sec:data}. 

\subsection{Lexical Constraints} Underspecification is a major challenge in VQG evaluation: given an image and an answer, there is often an intractably large set of questions that could have generated the answer. 
For example, in \cref{fig:fig1}, the answer ``purple'' could also correspond to the question, ``What color is the bottle's base?" 
Furthermore, even when the question is about the same topic, there are often a large number of semantically identical ways to phrase the question which may have very different surface forms.
This poses a problem for surface-level evaluation metrics like BLEU. 
Finally, in our task of rephrasing questions, similarity is not a perfect predictor of quality.
At one extreme, if the model generated the original question, it would receive a perfect similarity score when evaluated against the original question, but be as ambiguous as before.
At the other extreme, as illustrated in the preceding example, a model may generate a valid question conditioned on the answer that has no relation to the original question's intent. 

We attempt to tackle this problem by including positive lexical constraints from the original question in our decoding process.
In a normal VQG setting, this would be impossible, since it requires the question at test time. 
However, in our setting, where the goal is to \emph{rephrase} visual questions, we can assume access to questions. 
To generate a question on the same topic as the original, we use fast lexically-constrained decoding \citep{post.m.2018} with disjunctive positive constraints \citep{hu.e.2019} during test decoding (+c in \cref{tab:gen_results}). We extract all contiguous noun spans from the question using Spacy's part-of-speech tagger \citep{honnibal.m.2017}; these are added as disjunctive positive beam search constraints so that the output contains at least one span. For example, without constraints, the question ``Where are the people sitting?'' (answer: ``park'') is rewritten ``What kind of park is this?'', while with constraints the model predicts ``Where are the people?'' 

% \vspace{-0.5em}
\subsection{Baselines}
Due to the difference in our train and validation data as well as our use of constraints, our results are not directly comparable to previous VQG models. 
We instead compare our model to two baselines: ``no image'' (-v) and ``no answer'' (-t), where we give our model only the answer and only the image, respectively. 
These ablations verify our model's integration of multimodal information. 

% \vspace{-0.5em}
\subsection{Training}
We use the VQAv2 training set for training, excluding the examples we annotated, which came from the train split.
Since the answers for the VQA test split are not public, we use the validation data for testing and validation. 
We take $2,000$ questions pairs for validation and hold out the remaining $\sim 21K$ for testing. 
Each model was trained to convergence, measured by 5 consecutive epochs without BLEU score improvement, on four NVidia Quadro RTX 6000 GPUs; training took about 40 hours per model. 
All models were trained with the same hyperparameters (cf. \cref{append:hyper}). 

\section{Visual Question Generation} \label{sec:vqg_exp}
Before analyzing performance on our dataset, we verify that the question-generation model we proposed is able to generate reasonable questions for the dataset more broadly. 
Here, we follow past work in reporting several string-based metrics: BLEU \citep{papineni.k.2002}, CIDEr \citep{vedantam.r.2015}, Rouge-L \citep{lin.c.2004} scores.  
We also report BertScore \citep{zhang.t.2019}. 

\begin{table}[ht]
    \small
    \begin{tabular}{ccccc}
        \toprule
           Model &  BLEU-4 &  CIDEr &  ROUGE-L &  BERT \\
        \midrule
        iVQA$^*$ &    0.21 &   1.71 &     0.47 &   N/A \\
        \hline
        VT5-v &    0.22 &   1.51 &     0.45 &  0.93 \\
        VT5-v+c &    0.21 &   1.82 &     0.47 &  0.93 \\
        VT5-t &    0.16 &   1.00 &     0.32 &  0.92 \\
        VT5-t+c &    0.18 &   1.51 &     0.38 &  0.92 \\
           \hline
        VT5 &    0.27 &   1.98 &     0.48 &  0.94 \\
        VT5+c &    0.26 &   2.21 &     0.50 &  0.94 \\
        \bottomrule
    \end{tabular}
    % \vspace{-0.5em}
    \caption{Test performance of the VQG model and baselines. Our model is able to integrate multimodal information and produce high-similarity questions.}
    \vspace{-0.5em}
    \label{tab:gen_results}
\end{table}

\cref{tab:gen_results} shows the test performance of the models tested, with and without constrained decoding. 
We see that the proposed generation model outperforms both baselines by a wide margin, indicating that it is successfully integrating information from both modalities.
Furthermore, we see that in all cases, constraints improve performance; this is unsurprising, since the constraints force the model to include more of the reference question's n-grams. 
Finally, we include the performance of the iVQA model from \citet{liu.f.2018} in this table; however, we stress that the numbers are not directly comparable, since the training and evaluation data is different. 
Nevertheless, they help assert that our model is within the correct range for VQG.

% \vspace{-0.5em}
\subsection{Model as an Annotator}
In \cref{sec:data} we measured the inter-annotator agreement between annotators for clustering.  
We now compare the model predictions to these annotations with the same metric.
Specifically, we measure how well the model's answer clusters align with annotated clusters, assuming access to the number of clusters given by the annotators. 
While this is a limiting assumption, it lets us evaluate to what degree the model's representations are useful in grouping answers, independently of whether the clustering algorithm can infer the right number of clusters. 
We hypothesize that the VQG loss will result in answer representations for answers to the same underlying question being more similar than answer representations for different underlying questions. 

In order to obtain clusters from model representations, we use the \kmeans{} algorithm to group model representations of each answer $a_i \in A_i$.
We then compare the F1 overlap between clusters produced by the model (and different clustering baselines) to the clusters produced by annotators using the method detailed in \cref{sec:data}. 
We compare against several simple baselines. 
The \textbf{random} baseline randomly assigns answers to $K$ clusters.
The \textbf{perfect precision} baseline puts each answer in a separate cluster, leading to perfect precision but poor recall. 
The \textbf{perfect recall} baseline clusters all of the answers together, leading to perfect recall but poor precision.
We also take the initial clustering of GloVe vectors with \kmeans{}, using an incrementally increasing $K$, as described in \cref{sec:data}, as a baseline.
For a more direct comparison, we extract the frozen pre-trained ViLT representation for the answer tokens and use mean pooling to combine them into a single vector per answer, clustering them with \kmeans{} for the \textbf{ViLT+\kmeans{}} baseline. 
Note that the ViLT representation is frozen and not trained for VQG. 
This baseline is contrasted with the \textbf{VT5 + \kmeans{}} system, where we extract mean-pooled answer token representations from the final layer of our VQG encoder and use these for clustering with \kmeans{}. 
Gains over the ViLT baseline reflect the benefits of the VQG loss combined with the T5 encoder pre-training. 

\begin{table}[]
    \centering
    \begin{tabular}{lccc}
    \toprule
    Method & Avg. P & Avg. R & Avg. F1 \\
    \hline
    Human$^*$ & 88.6 & 91.7 & 88.4 \\
    \hline 
    Random & 64.9 & 70.4 & 59.4 \\
    Perfect P & 100.0 & 50.6 & 61.1 \\
    Perfect R & 63.4 & 100.0 & 76.3 \\
    GloVe initial & 98.4 & 64.3 & 72.4 \\
    ViLT + \kmeans{} & 65.9 & 68.6 & 60.1 \\  
    \hline
    VT5 + \kmeans{} & 81.9 & 84.0 & \textbf{79.0} \\
    \bottomrule
    \end{tabular}
    % \vspace{-0.5em}
    \caption{Clustering metrics; Human results included for indirect comparison only.}
    \vspace{-0.5em}
    \label{tab:clust}
\end{table}

\cref{tab:clust} shows the clustering results. 
We see that VT5+\kmeans{} outperforms all baselines in F1, indicating that the representations learned via a VQG objective contain answer-group information.
This is surprising, as the objective here does not directly optimize for answer groups; for a given training example ($I_i, a_i, Q_i$), there is a single reference output $Q_i$ for all answers, regardless of the group they are in. 
However, the grouping information might be found in the dataset more broadly; when considering multiple examples with similar answers, answers in the same group may correspond to similar questions, leading them to be closer in representation space and thus in the same \kmeans{} cluster. 
In other words, the encoder representation for a given answer, having been trained across many similar questions and answers, is more similar within an answer group than across groups. 

\section{Human Evaluation} \label{sec:human}
The metrics in \cref{sec:vqg_exp} suggest that our model holds promise as a method for rephrasing ambiguous questions; \cref{tab:gen_results} indicates that the model produces fluent questions conditioned on images and answers, and \cref{tab:clust} shows that the model contains some of the requisite information for rewriting questions according to the answer clusters from human annotators. 
However, these automated metrics fall short of providing a full picture of the quality of rewritten questions, especially because, as mentioned before, it is not clear that similarity is a monotonic measure of success in our case. 
Thus, we conduct a human evaluation of $100$ rewritten questions, specifically testing whether rephrased questions (from annotators and from the model) are less ambiguous than their original counterparts from the VQA dataset. 

\vspace{-0.5em}
\subsection{Methods}
\vspace{-0.5em}
Our evaluation paradigm presents annotators with an 3-way ordinal decision (``yes'', ``maybe'', ``no''), rating whether an answer is appropriate given an image and question.
We sample 100 examples from our dataset; each example is paired with 3 questions: annotator-generated, model-generated, and original (from the VQAv2 dataset).
The model-generated questions are taken from the VT5 model with constraints. 
For each image-question pair, we obtain 2 answers -- one from the answer group corresponding to the rewritten question, and a distractor answer from a different answer group, as determined by the human annotations.
In other words, for the example from \cref{fig:fig1}, one non-distractor instance in the evaluation HIT would be the image, the question ``What species of flowers are these?", and the answer ``daisy'', while the distractor instance would have the answer ``purple''. We would also have these two answers paired with the question ``What kind of flowers are these?". 
An ambiguous question should be rated as acceptable for \emph{both} answers (the actual and distractor), while a question rephrased to be less ambiguous should be rated as acceptable for the actual answer but \emph{not} for the distractor answer, which corresponds to a different underlying question. 
Annotators were paid $0.04$ per annotation for a total of 600 annotations, or $\sim \$16$ per hour, and did not participate in the main annotation task. 

% \vspace{-0.5em}
\subsection{Results and Analysis} 
\begin{figure}
    \centering
    \includegraphics[width=0.49\textwidth]{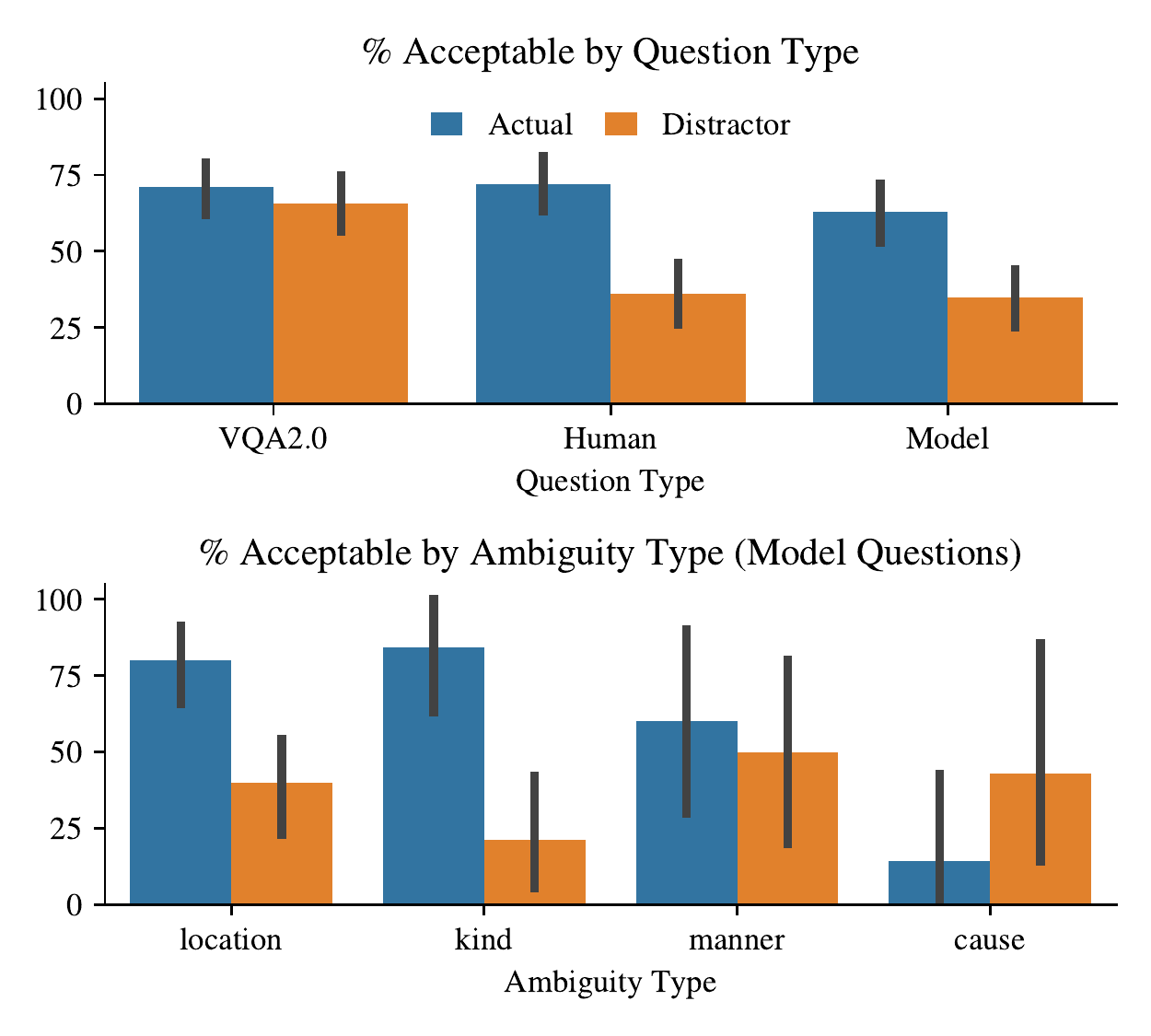}
    \vspace{-2.5em}
    \caption{Percentage of answers rated as acceptable for each question type (annotator-rewritten, model-rewritten, original). Error bars represent bootstrapped $95\%$ confidence intervals. Rewritten questions are less ambiguous (distractor rated as unacceptable) than their original counterparts. Model questions are generally less ambiguous across ambiguity categories.}
    \vspace{-1em}
    \label{fig:human_eval}
\end{figure}

\cref{fig:human_eval} shows the percentage of answers rated as acceptable (``yes'' as opposed to ``maybe'' and ``no'') across different conditions. 
The original, unedited question shows no significant difference between the actual and distractor answer, as measured by McNemar's test \citep{mcnemar.q.1947}. 
This is expected, given that both answers (e.g. ``daisy'' and ``purple'') were given by annotators in the original dataset to the original question, and thus are both likely to be viewed as acceptable. 
Both types of edited questions, on the other hand, show a significant difference between the actual answer and distractor answer, indicating that questions rephrased by annotators and by the model more specifically select answers from one answer group over, i.e. they are less ambiguous with respect to the answer group.
The fact that the questions predicted by the model show only a small drop is promising, as it indicates that the model outputs are fluent and faithful to the original topic. 
Nevertheless, the model's questions are rated as slightly less acceptable than the human questions, indicating room for improvement.
In the bottom of \cref{fig:human_eval} we see the percentage broken out by ambiguity type for the four most frequent types; here, we plot only the model-predicted sentences. We see that across most types there is a drop, with model outputs being rated as acceptable with the true answer, but not with the distractor. 

% \vspace{-0.5em}
\section{Discussion}

While there are many competing accounts of question semantics, a general consensus maintains that the analysis of questions requires an analysis of the dialogic context in which they exist \citep{ginzburg.j.2010}. 
This differs substantially from the VQA domain, where questions are being posed and answered outside of a dialogue. 
Specifically, annotators are asked to answer questions in a single attempt, with no recourse to follow-up questions or interactions. 
One umbrella reason for disagreement and ambiguity in this setting that we have not explored in our analysis is guided by the Question-Under-Discussion (QUD) framework \citep{roberts.c.1996}, which frames a dialogue as a series of moves around asking and resolving QUDs.
In the setting we explore, much of the disagreement seems to be driven by annotators disagreeing on which QUD to resolve with their response to the question; it is possible that having additional context would reduce the disagreement between annotators by providing additional information.
Indeed, the presence of contextual information is a key reason given by \citet{piantadosi.s.2012} for the existence of ambiguity in communication systems like natural language: if a listener can reasonably be expected to reconstruct a speaker's intended meaning from the speaker's utterance combined with the prior context, then the speaker does not have to specify the (redundant) contextual information in their utterance, lowering their effort. 
Given the artificial nature of the VQA annotation setting, with one set of annotators producing questions and another answering them in a single turn, it is unsurprising that ambiguities arise. 

Similarly, the rewritten questions from our dataset and our rewriting model can be framed as questions specifying which QUD is being resolved by a given answer. 
This is especially true of property-based questions, where the QUD influences which property is used to partition the space of possible answers (cf. \cref{sec:groenendijk}). 

\vspace{-0.5em}
\subsection{Limitations}  Our primary limitation is the size of our collected dataset; we have collected a quality dataset which we demonstrated is useful for analysis, but which is too small for training large-scale neural models. 
However, \cref{sec:vqg_exp} indicates that a training-size dataset may not be necessary, as our question generation model is capable of capturing answer groups without explicit supervision. 
Another limitation on our dataset is the relative subjectivity of the task; in completing the annotation, we found that identifying ambiguity and isolating the different underlying questions often involves a Gestalt shift. 
Once an interpretation of the question is chosen, it becomes increasingly hard to see any other.
This makes the annotation task subjective; where one annotator might see ambiguity leading to multiple valid answers, another might see one correct answer group and a number of invalid ones. 
Thus, the annotations in our dataset represent a high precision subset (rather than a high-recall subset) of all the possible ambiguous datapoints. 
This subjectivity also risks introducing annotator bias (including the author's own biases) into the data; we acknowledge that the vetting steps by the authors may have compounded this further. 
We are also limited by the quality of the underlying data. 
Our dataset builds on the VQAv2 dataset \citep{goyal.y.2017} and the annotations from \citet{bhattacharya.n.2019}, both of which were large-scale annotation efforts intended for training.
Due to their scale, individual datapoint quality is often  quite low; this was one factor contributing to the need for post-hoc cleaning in the annotation process. 

\subsection{Future Work} 
In addition to addressing these limitations, we leave exploiting the rewriting model to future work.
In \cref{tab:gen_results} and \cref{fig:human_eval} we demonstrated that our question rephrasing model works well for producing fluent questions that reduce ambiguity. 
Furthermore, in \cref{tab:clust} we showed that the model's representations contain information about the underlying question being asked, even though this information is not directly present in the training data and we do not include any supervision from our dataset. 
Future work could examine utilizing the rephrasing model in a search-engine environment, where users are actively querying about images. 
Given an ambiguous question identified and a set of answers to it from a VQA model, our model could be used to rephrase the question according to each answer. 
Just as a presenter will often rephrase a question from the audience, the model might present the user with the rephrased question it is actually answering, which would result in better interpretability. 
This improved interpretability might teach users how to interact with the model. 

\section{Related Work} \label{sec:related}

\subsection{Ambiguity} Ambiguity in question-answering has been explored in the past: \citet{min.s.2020} introduce AmbigQA, a dataset of ambiguous open-domain questions paired with disambiguated rewrites. 
Our dataset differs in its domain: we address visual questions. 
Additionally, many of the ambiguities in AmbigQA are a result of background knowledge and changing dynamics.
This is further explored by \citet{zhang.m.2021}, who introduce SituatedQA, a dataset of context-dependent questions and answers. 
In contrast, because VQA questions are closed-domain (i.e. they are typically about an image, not the world in general) the ambiguities we explore are more often a result of the language used in the question, rather than background knowledge of the annotator. 
Ambiguity has also been explored in natural language inference (NLI): \citet{pavlick.e.2019} explore annotator disagreement on NLI examples, finding ambiguity to be one source of disagreement. 

% \vspace{-0.5em}
\subsection{Disagreement in VQA} 
After the introduction of VQA datasets such as VQAv2 \citep{goyal.y.2017} and VizWiz \citep{gurari.d.2018}, several papers focused on describing and diagnosing annotator disagreement in VQA.  
One line of work with deep ties to ours focuses on modeling annotator disagreement. 
\citet{gurari.d.2017} and \citet{yang.c.2018} present models for predicting annotator disagreement, which they use to reduce annotation cost. 
They both offer preliminary explorations of the features of high-disagreement questions. 
\citet{bhattacharya.n.2019} explore the reasons for disagreement in greater depth, annotating $\sim 45,000$ examples for the reason of disagreement; one of the possible reasons for disagreement is ambiguity.
We use these in our collection (cf. \cref{sec:data}).
However, the data labelled as ambiguous in \citet{bhattacharya.n.2019} covers a range of phenomena, including visual ambiguity and underspecification, whereas our focus is specifically on linguistic ambiguity in visual questions. 

% \vspace{-0.5em}
\subsection{Visual Question Generation}
Our work also relates to visual question generation (VQG).
While VQG was first introduced as a task of generating unconstrained questions about images \citep{mora.i.2016, mostafazadeh.n.2016}, subsequent work has explored conditioning on images and answers to produce questions, as in \citet{liu.f.2018}.
\citet{li.y.2018} propose to generate questions as a dual auxiliary task for VQA, and \citet{shah.m.2019} use cycle consistency between generation and answering for improving VQA. 
Some past work has conditioned on partial answer information: \citet{krishna.r.2019} condition on answer categories rather than full answers, and \citet{vedd.n.2022} present a latent variable model which allows answers to be imputed at test-time. 
\citet{terao.k.2020} condition on answer-distribution entropy; in a similar vein to our work, \citeauthor{terao.k.2020} focus on VQG for ambiguous questions. 
However, \citeauthor{terao.k.2020} define ambiguity according to the entropy of their trained model and rely on user-specified entropy values for inference; we define it in a model agnostic way, according to features of the input.
They also do not distinguish between linguistic and visual ambiguity.

% \vspace{-0.5em}
\section{Conclusion}
% \vspace{-0.5em}
We have presented a dataset of ambiguous VQA questions, annotated with reasons why they are ambiguous, as well as answers grouped by the underlying disambiguated question they are answering. 
We then introduced a model for rephrasing ambiguous questions according to their answers, finding that the model, which is trained purely on visual question generation, is able to recover information about the underlying question. 
We validate both our dataset and model using automatic and human evaluations, where we find that both reduce question ambiguity.

\section*{Acknowledgements} 
We would like to thank the reviewers for their helpful comments and feedback, especially Reviewer 1, who suggested we include a broader discussion of QUDs and provided several helpful pointers. 
We would also like to thank Nils Holzenberger and Kate Sanders for feedback on earlier drafts. 
This work was funded in part by NSF $\#1749025$ and an NSF GRFP to the first author.
\bibliography{vqa}
\bibliographystyle{acl_natbib}

\appendix

\label{sec:appendix}
\input{appendix}

\end{document}

%% file: ontology_table.tex
\begin{table*}[ht]
    \centering
    \small
    \begin{tabular}{p{0.09\linewidth} | p{0.09\linewidth} |p{0.05\linewidth}|p{0.58\linewidth} | p{0.04\linewidth}}
    \hline
    Category & Property & PropB. & Description & Ex. \\
    \hline
    \multirow{3}{0.09\linewidth}{Property-based} & \cellcolor{lg} Location & \cellcolor{lg} \texttt{LOC}  & \cellcolor{lg} Asks about an object's location. & \cellcolor{lg} \ref{sec:onto_loc} \\\cline{2-5}
     &  Time  & \texttt{TMP}  &  Asks about the time of an event or the time a picture was taken. &  \ref{sec:onto_time} \\\cline{2-5}
     & \cellcolor{lg} Kind & \cellcolor{lg} N/A  & \cellcolor{lg} Ask about what kind of something an object is. & \cellcolor{lg} \ref{sec:onto_kind} \\
     \hline 
    \multirow{5}{0.09\linewidth}{Dynamic} & Cause & \texttt{CAU}  & Ask for the cause of an event. & \ref{sec:onto_cause}\\\cline{2-5}
    & \cellcolor{lg} Purpose  & \cellcolor{lg} \texttt{PRP}  & \cellcolor{lg} Ask for the purpose of an event. & \cellcolor{lg} \ref{sec:onto_purpose} \\\cline{2-5}
    & Goal & \texttt{GOL}  & Ask for the goal (location or person) of an object or event. & \ref{sec:onto_goal_direction} \\\cline{2-5}
    & \cellcolor{lg} Direction & \cellcolor{lg} \texttt{DIR}  & \cellcolor{lg} Ask for the path being taken by an object.  & \cellcolor{lg} \ref{sec:onto_goal_direction} \\\cline{2-5}
    & Manner & \texttt{MNR}  & Ask in what manner an event is happening. & \ref{sec:onto_manner} \\
    \hline 
    \multirow{4}{0.09\linewidth}{Pragmatic and Other} & \cellcolor{lg} Multiple  & \cellcolor{lg} N/A  & \cellcolor{lg} Ask annotators to choose one of multiple options. & \cellcolor{lg} \ref{sec:onto_multiple}  \\\cline{2-5}
    & Grouping & N/A  & Ask annotators to group multiple items. & \ref{sec:onto_grouping} \\\cline{2-5}
    & \cellcolor{lg} Uncertainty & \cellcolor{lg} N/A  & \cellcolor{lg} Contain visual uncertainty, especially for questions about events. & \cellcolor{lg} \ref{sec:onto_uncertainty} \\\cline{2-5}
    & Mistake & N/A & These involve bad answers or bad questions/images. & \ref{sec:onto_bad_annotator} \\
     \hline
    \end{tabular}
    \vspace{-0.5em}
    \caption{Ontology of reasons why examples are ambiguous. Examples and details in \cref{append:ontology}.}
    \vspace{-1em}
    \label{tab:ontology}
\end{table*}
\label{ref:big_table}

%% file: appendix.tex
\newpage
\section{Crowdsourcing} \label{append:crowd}
To collect a set of vetted data, a pilot task (or HIT) was run.
A local annotator was paid $\$15$ for one hour of annotation time (including watching the instruction video). 
The same annotations were then annotated by one of the authors.
During this phase, the authors themselves ensured that there was no personally identifiable or offensive material in the data.
From this data, we generated a set of examples for a pilot HIT to be run on Amazon's MechanicalTurk (MTurk).

To identify high-quality MTurk annotators, we ran pilot HIT of 41 examples from the local annotations, with 28 examples marked as ambiguous in the pilot and 13 examples marked as unambiguous (e.g. skipped).
Workers were restricted to be located in the US.
The annotations were presented sequentially, so that annotators had to complete all 41 examples to complete the HIT. 
Annotators were paid $\$0.10$ per example and received a $100\%$ bonus for completing all examples ($\$8$ per HIT, roughly $\$16$ per hour of annotation). 

From the pool of MTurk annotators who completed the pilot, we identified the top  annotators. 
We then presented them with 850 examples in a non-sequential format, where each annotator could do as many as desired. 
No examples were flagged as offensive in this stage.
Two annotators completed the task, which paid $\$0.10$ per example, with an $\$8$ bonus for every 300 examples.
This corresponded to roughly $\$16$ per hour. 

\section{VQA Ambiguity Ontology} \label{append:ontology}
\subsection{Question Semantics} \label{sec:groenendijk}
Formal semantics often focuses on variants of truth-conditional semantics, where knowing the meaning of an utterance is equated to knowing the conditions that would make the utterance true \citep{davidson.d.1967}. 
This account handles propositions well; however, evaluating the truth conditions of questions, an equally central feature of human language, seems more challenging. 
A rich literature has explored the meaning of questions \citep[i.a.]{hamblin.c.1958, belnap.n.1976, groenendijk.j.1984}; for the purposes of this overview, we will briefly touch on one proposal which is of particular relevance to several categories outlined in \cref{sec:data_ontology}. 
Under the partition semantics proposed by \citet{groenendijk.j.1984}, the meaning of a question is a set of utterances which partition the set of possible worlds.
This is best illustrated with an example: assuming there were only two people in the whole universe (``John'' and ``Mary''), then the meaning of the question ``Who walks?'' is the partition induced by the propositions ``Only John walks'', ``Only Mary walks'', ``Both walk'', ``Neither walks''. 
Each cell in the partition contains all possible worlds where the proposition is true, i.e. the ``John walks'' cell might contain a world where he walks outside, or on a treadmill, or one where the moon is made of cheese. 

This proposal will describe a core feature of one type of disagreement we find. 
In certain cases, different answerers may have a different set of propositions in mind, leading to incompatible partitions. 
For example, given a picture of a blue children's tshirt, the question, ``What kind of shirt is this'' might be answered with ``blue'', ``child's'', or ``small''. 
In each of these cases, the partition function may be different, i.e. the ``blue'' answer is given as opposed to other colors, while the answer ``child's'' stands against ``adult''.
In these cases, different partition functions define different sets of alternatives, leading to a variety of answers. 

\subsection{Property-based}
\begin{figure}[H]
    \centering
    \includegraphics[width=0.3\textwidth]{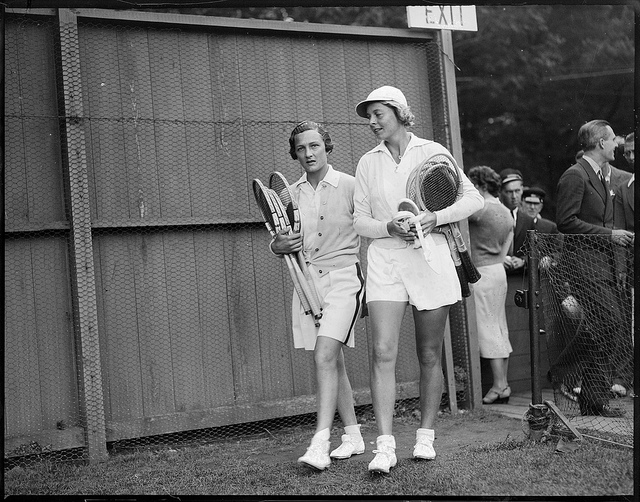}
    \caption{Question: What kind of top is she wearing? Answers: ``white''; ``button up to''; ``sweater''; ``long sleeve''}
    \label{fig:kind_example}
\end{figure}

Property-based ambiguities stem from annotators choosing to report different properties of objects or events with multiple properties. 
% Another way to think of property-based ambiguities is in terms of the partition-based question semantics of \citet{groenendijk.j.1984}.
% Under partition semantics, the meaning of a question is a partition over possible worlds. 
% These partitions can be described in terms of equivalence classes; for example, given a universe of two people (``John'', ``Mary'') the partition induced by the question ``Who walks?" has 4 cells, containing all worlds where only John walks, only Mary walks, both walk, and neither walk. 
% In property-based ambiguities, annotators seem to choose different equivalence classes, which correspond to different cells in a partition and different sets of alternatives.  
As mentioned, these relate closely to \citet{groenendijk.j.1984}'s question semantics, where the meaning of a question is defined as a partition over possible worlds, and where different meanings would result in different partitions. 
For example, in \cref{fig:kind_example}, the annotator who says ``white'' is partitioning according to colors (e.g. ``white sweater'' as opposed to ``blue sweater'' or ``black sweater'') while the annotator who says ``long sleeve'' is partitioning possible worlds according sleeve style. 
% Crucially, these partitions are defined by equivalence classes, meaning that the property used to define the partition (i.e. color, sleeve length, etc.) is the salient feature, and items differing along other features are treated as equivalent. 
% Thus, if the equivalence class is defined by color, then all blue shirts, whether short or long-sleeved, would be in the same cell of the partition. 

% \subsection{Property-based}
There are three sub-classes of property-based ambiguities: location, kind, and time.  

\subsubsection{Location} \label{sec:onto_loc}
Location maps to the PropBank tag {\tt{ARGM-LOC}}.  Answers here typically differ in terms of frame-of-reference, tracking with the observations of \citet{viethen.j.2008}. 
\hyperref[ref:big_table]{(Back to table)}
\begin{figure}[H]
    \centering
    \includegraphics[width=0.3\textwidth]{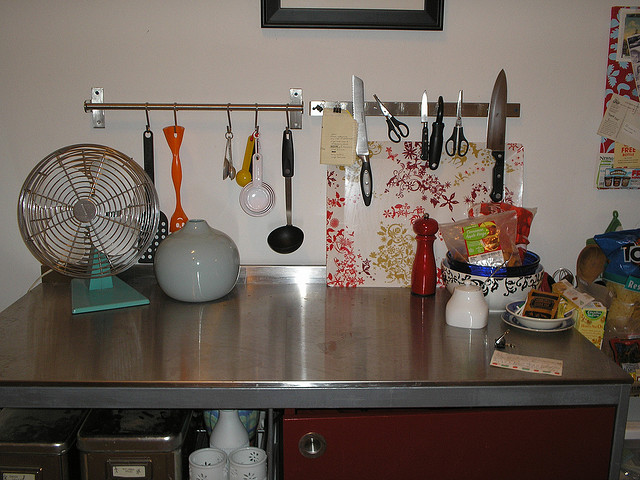}
    \caption{Question: Where is the fan? Answers: ``on table''; ``[l]eft side of counter in kitchen''}
    \label{fig:loc_example}
\end{figure} 

\subsubsection{Time} \label{sec:onto_time}
This category maps to the PropBank tag {\tt{ARGM-TMP}}.  Answers often differ in terms of granularity and frame-of-reference (e.g. ``morning'', ``breakfast time'', ``8am'').
\hyperref[ref:big_table]{(Back to table)}
\begin{figure}[H]
    \centering
    \includegraphics[width=0.3\textwidth]{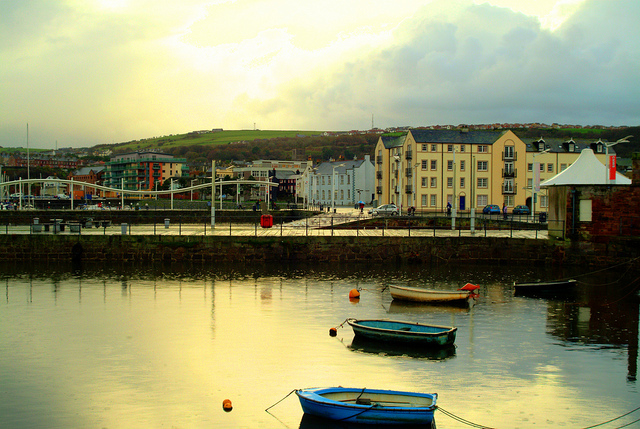}
    \caption{Question: What time of day is it? Answers: ``morning''; ``4 o'clock''}
    \label{fig:time_example}
\end{figure}

\subsubsection{Kind} \label{sec:onto_kind}
These do not map to PropBank, and ask about what type or kind of something an object is. Answers differ in terms of property class chosen. 
\hyperref[ref:big_table]{(Back to table)}

\subsection{Dynamic} 
Dynamic questions are typically about properties of dynamic objects or events. 
Annotators often disagree on the type of question being asked (e.g. \emph{cause} vs. \emph{purpose}), as well as the underlying question within a type. 
These questions commonly correspond to ``why'' and ``how'' questions.

\subsubsection{Cause} \label{sec:onto_cause} Maps to \texttt{ARGM-CAU}. These ask for the cause of an event. Since cause and purpose are often ambiguous \citep{chapman.c.2016} annotators may differ here, and since cause is often under-specified from a static image, annotators may impute different causes. 
Even when causes are not imputed, annotators often may choose one of multiple causes, or report causes at different levels of granularity. 
\hyperref[ref:big_table]{(Back to table)}
\begin{figure}[H]
    \centering
    \includegraphics[width=0.3\textwidth]{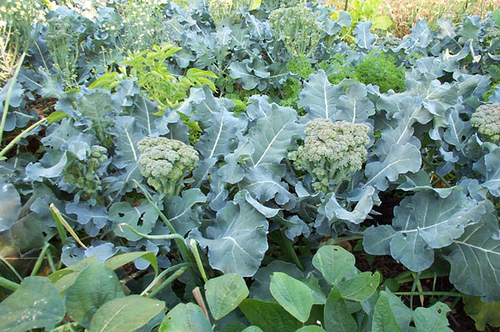} 
    \caption{Question: Why is this blue and green? Answers: ``it's vegetables''; ``cold''; ``photosynthesis''; ``garden''}
    \label{fig:cause_example}
\end{figure}

\subsubsection{Purpose} \label{sec:onto_purpose} maps to {\tt{ARGM-PRP}}. Purpose questions ask for the purpose of an event, and share their features with the \emph{cause} examples. 
\hyperref[ref:big_table]{(Back to table)}
\begin{figure}[H]
    \centering
    \includegraphics[width=0.3\textwidth]{figures/ambiguity_purpose.jpg} 
    \caption{Question: What is the netting for? Answers:  ``baseball''; ``ball''; ``protect public''; ``protect spectators''; ``safety''; ``don't get hit by ball'';}
    \label{fig:purpose_example}
\end{figure}

\subsubsection{Goal and Direction} \label{sec:onto_goal_direction}
Goal maps to \texttt{ARGM-GOL} and asks for the eventual goal (location or person) of an object or event. 
When the goal is a person, it is often the person who benefits from an action. 
Goals are often imputed, and can often be ambiguous with direction. 
Direction maps to \texttt{ARGM-DIR} and asks for the path being taken by an object. This is often ambiguous with goal, and is also often imputed or dependent on the frame-of-reference. 
\hyperref[ref:big_table]{(Back to table)}
\begin{figure}[H]
    \centering
    \includegraphics[width=0.3\textwidth]{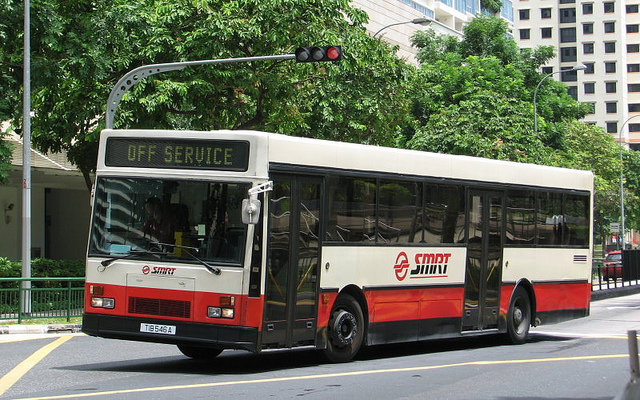}
    \caption{Question: Where is the bus going? Answers: ``station''; ``around corner''}
    \label{fig:goal_example}
\end{figure}

\subsubsection{Manner} \label{sec:onto_manner}
Manner maps to {\tt{ARGM-MNR}} and asks in what manner an event is happening. 
Manner questions can be ambiguous with cause questions. 
\hyperref[ref:big_table]{(Back to table)}
\begin{figure}[H]
    \centering
    \includegraphics[width=0.3\textwidth]{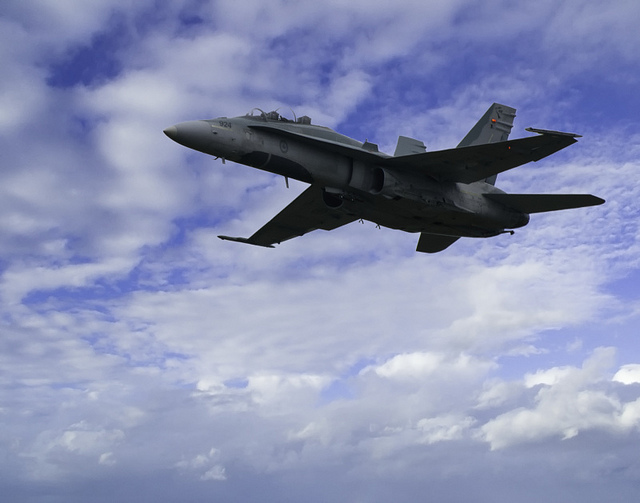}
    \caption{Question: How is the plane flying? Answers: ``low''; ``engines''; ``in air''}
    \label{fig:manner_example}
\end{figure}

\subsection{Pragmatic/Other}
Pragmatic ambiguities are typically characterized by an underspecified question which requires the answerer to infer a preference on the part of the questioner. 
For example, in the ``Multiple Options'' ambiguity, there are several valid responses, and different answerers might infer that different options are more or less salient to the questioner. 
None of the pragmatic ambiguities are aligned with PropBank.

\subsubsection{Multiple Options} \label{sec:onto_multiple}
A common source of disagreement is when annotators are asked to choose one of multiple options. For example, a question like ``what color is X?'' when X has multiple colors will often result in a variety of answers. 
Here, the ambiguity is with respect to the inferred intent of the questioner; the answerer must infer which option is most salient to the questioner.  
\hyperref[ref:big_table]{(Back to table)}
\begin{figure}[H]
    \centering
    \includegraphics[width=0.3\textwidth]{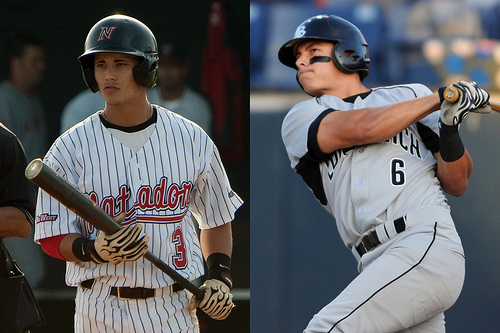}
    \caption{Multiple options ambiguity example. Question: What team is the man holding the bat playing for? Answers: ``matadors''; ``yankees''}
    \label{fig:multiple_example1}
\end{figure}

\subsubsection{Grouping} \label{sec:onto_grouping}
Grouping ambiguity often co-occurs with multiple options, and involves grouping several options; different annotators may include or exclude items from their groups.  
\hyperref[ref:big_table]{(Back to table)}
\begin{figure}[H]
    \centering
    \includegraphics[width=0.3\textwidth]{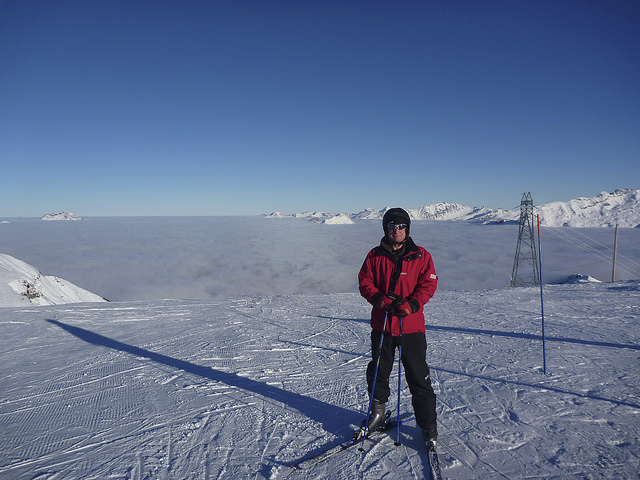}
    \caption{Question: What is on the right of the picture? Answers: ``sky posts''; ``mountain''; ``electric tower, ski pole, and mountain top'';} 
    \label{fig:grouping_example}
\end{figure}

\subsubsection{Uncertainty} \label{sec:onto_uncertainty}
 Many examples contain visual uncertainty, especially for questions about events, which are inherently hard to capture in a static image. 
 \hyperref[ref:big_table]{(Back to table)}
\begin{figure}[H]
    \centering
    \includegraphics[width=0.3\textwidth]{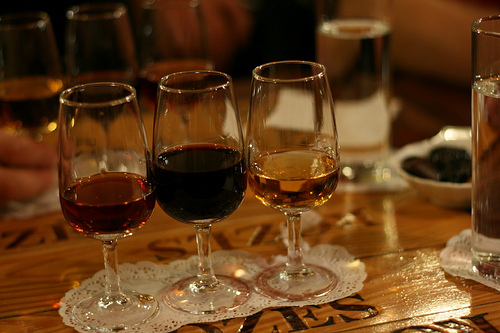}
    \caption{Uncertainty example. Question: Where is the white wine bottle? Answers: ``not visible''; ``empty''}
    \label{fig:uncertainty_example}
\end{figure}

\subsubsection{Annotator mistakes} \label{sec:onto_bad_annotator}
Some annotators provide bad or unreasonable answers to questions. 
\hyperref[ref:big_table]{(Back to table)}
\begin{figure}[H]
    \centering
    \includegraphics[width=0.3\textwidth]{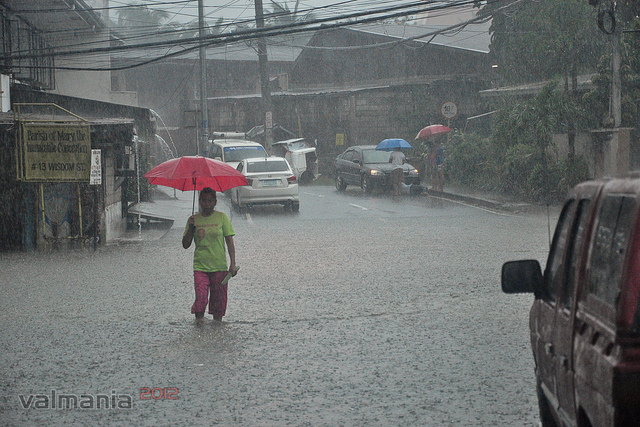}
    \caption{Annotator mistake. Question: How high is the water? Answers: ``2-3 inches''; ``rain water''}
    \label{fig:ma_example}
\end{figure}

\subsubsection{Bad question/bad image} \label{sec:onto_bad_image}
Some questions are nonsensical and some images are extremely low quality, making answering any question about them impossible. 
\hyperref[ref:big_table]{(Back to table)}
\begin{figure}[H]
    \centering
    \includegraphics[width=0.3\textwidth]{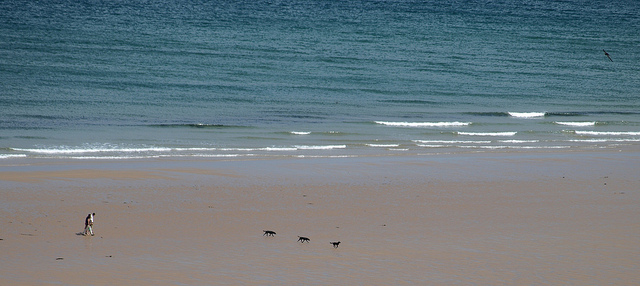}
    \caption{Bad image or data. Question: Which bird looks about to take off the ground? Answers: ``middle bird''; ``left 1''}
    \label{fig:mb_example}
\end{figure}

\section{Interface} \label{append:interface}
\begin{figure*}
    \centering
    \includegraphics[width=\textwidth]{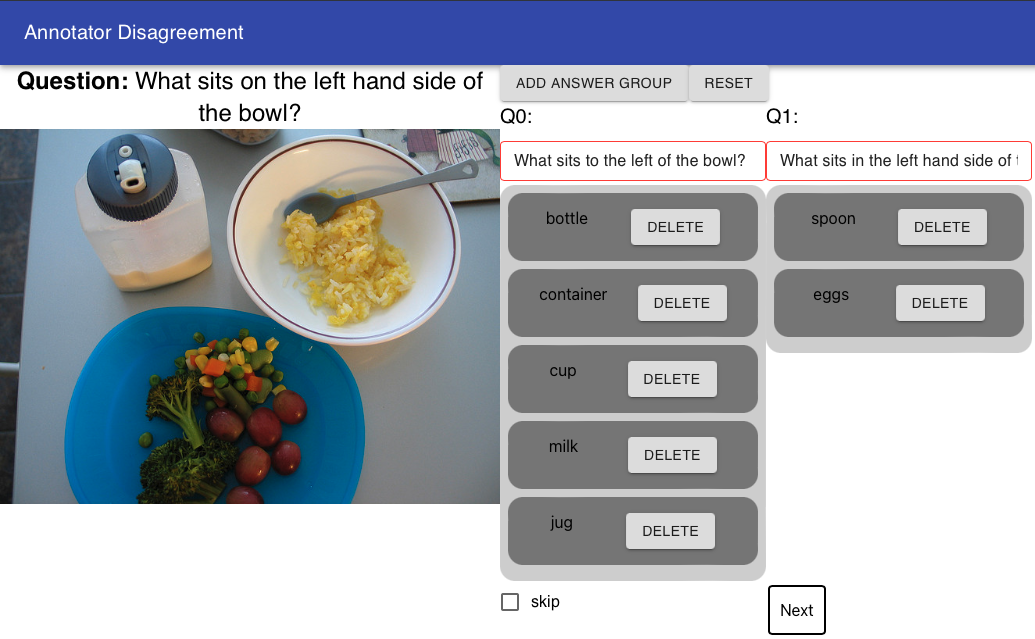}
    \caption{The annotation interface. }
    \label{fig:interface}
\end{figure*}

\cref{fig:interface} shows the annotation interface used to collect the dataset. Answers are drag-able objects and can be moved across columns. New answer groups can be added. Questions are auto-populated with the original question and then edited by the annotator. Skipping opens up a text box with an auto-populated reason (``All answers to the same question'') that can be edited.

\section{Hyperparameters} \label{append:hyper}
Models were trained with the AdamW optimizer \citep{loshchilov.i.2018} using a learn rate of \num{1e-4} with linear weight decay of $0.01$.
The learn rate followed a linear warmup schedule with $4,000$ warmup steps. 
The batch size was set to 32 per GPU, leading to an effective batch size of 128. 
As fine-tuning ViLT for VQG had no substantial impact, we freeze the ViLT encoder during training.

\section{Validation Performance}
\cref{tab:val_gen_results} shows the validation performance for all metrics reported in \cref{tab:gen_results}. Trends mirror those seen in the test data. 
\begin{table*}[]
    \centering
\begin{tabular}{lcccccccc}
\toprule
   Model &  BLEU-1 &  BLEU-2 &  BLEU-3 &  BLEU-4 &  CIDEr &  ROUGE-L &  METEOR &  BERT \\
\midrule
iVQA$^*$ &    0.43 &    0.33 &    0.26 &    0.21 &   1.71 &     0.47 &    0.21 &   N/A \\
\hline
   VT5-v &    0.47 &    0.31 &    0.22 &    0.16 &   1.05 &     0.42 &    0.41 &  0.93 \\
   VT5-t &    0.39 &    0.21 &    0.14 &    0.10 &   0.48 &     0.29 &    0.30 &  0.91 \\
     VT5 &    0.53 &    0.37 &    0.28 &    0.22 &   1.51 &     0.46 &    0.47 &  0.94 \\
\hline
 VT5-v+c &    0.47 &    0.30 &    0.21 &    0.15 &   1.33 &     0.43 &    0.45 &  0.93 \\
 VT5-t+c &    0.42 &    0.25 &    0.17 &    0.12 &   0.95 &     0.34 &    0.38 &  0.92 \\
   VT5+c &    0.53 &    0.37 &    0.27 &    0.21 &   1.73 &     0.47 &    0.50 &  0.94 \\
\bottomrule
\end{tabular}
\caption{Validation performance of the VQG model and baselines. Our model is able to integrate visual and textual information and output questions with high similarity to reference questions.}
\label{tab:val_gen_results}
\end{table*}

% \section{Why Ambiguity} \label{append:why}
% \cref{fig:why_examples} gives an example of each possible combination of dynamicity and agency. Answers are grouped by whether they are cause or purpose answers. 
% Note that grouping these answers is an underspecified task -- in many cases, both readings can be coerced using context. 
% \begin{figure}[H]
%     \centering
%     \includegraphics[width=0.5\textwidth]{figures/dynamic_agentive_examples.pdf}
%     \caption{Example of an ambiguous annotation for each category in our ``why'' question analysis }
%     \label{fig:why_examples}
% \end{figure}

\section{License} Code and data will be released under an MIT license.
% \begin{figure}[H]
%     \centering
%     \includegraphics[width=0.5\textwidth]{figures/necker.pdf}
%     \caption{Both squares in the Necker cube can be seen as the front of the cube, but not at the same time.}
%     \label{fig:necker}
% \end{figure}